\pdfoutput=1

\documentclass[11pt]{article}

\usepackage[final]{acl}

\usepackage{times}
\usepackage{latexsym}
\usepackage{amsmath}
\usepackage[T1]{fontenc}

\usepackage[utf8]{inputenc}

\usepackage{microtype}

\usepackage{inconsolata}

\usepackage{graphicx}

\usepackage{hyperref}
\usepackage{graphicx}
\usepackage{wrapfig}
\usepackage{url}
\usepackage{xcolor}
\usepackage{booktabs}
\usepackage{pgfplots}
\usepackage{threeparttable}
\usepackage{subcaption}
\usepackage{cleveref}
\usepackage{color,soul}
\usepackage{svg}
\usepackage{stackengine}
\usepackage{mdframed}
\usepackage{colortbl}
\usepackage{amsfonts}

\definecolor{Gray}{gray}{0.9}

\definecolor{LightCyan}{rgb}{0.75,1,1}

\newcommand{\method}{\textsc{kv-distill}\,}
\newcommand{\kv}{\textsc{kv}\,}

\title{\textsc{kv-distill}: Nearly Lossless Learnable Context Compression for LLMs}

\author{
Vivek Chari$^{1}$\ , Guanghui Qin$^{2}$, Benjamin Van Durme$^{1,2}$ \\
$^{1}$Johns Hopkins University \quad $^{2}$Microsoft \\
\texttt{\{vchari2,vandurme\}@jhu.edu} 
}

\begin{document}
\maketitle
\begin{abstract}
Sequence-to-sequence tasks often benefit from long contexts, but the quadratic complexity of self-attention in standard Transformers renders this non-trivial. During generation, temporary representations -- stored in the so-called \kv cache -- account for a large portion of GPU memory usage and scale linearly with context length. We introduce \method, a Transformer compression framework that distills long context \kv caches into significantly shorter representations in a \textit{question-independent} fashion. \method can be trained as a parameter-efficient adaptor for pre-trained models, and enables the compression of arbitrary spans of a context while {preserving pre-trained model capabilities}. We treat a compressed-uncompressed cache as a student-teacher pairing and apply a KL-type divergence to match the generated outputs. \method outperforms other compression techniques in worst-case extractive tasks and approaches uncompressed performance in long context question answering and summarization, and it can be fine-tuned on domain-specific contexts to reduce lengths by up to 99\% while preserving downstream performance. We demonstrate the generalizability of \method across various model sizes and architectures.\footnote{Our code and checkpoints will be made available soon at this \href{https://github.com/vnchari/kv-distill}{link}}
\end{abstract}
\section{Introduction}

Harnessing the full potential of attention-based large language models (LLMs) often requires them to condition on long contexts. However, use of expansive contexts is complicated by the quadratic complexity of self-attention. In particular, during generation, one must maintain a store of all past key and value representations of past tokens (called the \kv cache) that grows linearly with sequence length. The memory burden imposed by the \kv cache is significant, and often limits the length of the sequences that a model can handle. 

Much work has been devoted to architectural improvements to  attention in order to reduce memory during generation. Strategies include augmenting sequences with memory tokens \citep{Rae2019-jo, Wu2022-qi}, sparsifying attention patterns \citep{Beltagy2020-dv}, and using conditional computation to only process essential tokens \citep{Ainslie2023-fy}. However, such techniques have seen little widespread adoption due to performance drops on downstream tasks, or inefficient training/inference procedures. Even when given long contexts without compression, LLMs fail to fully utilize them \citep{246867127,Liu2023-ne, Lu2024-hp}. Together this suggests long contexts may allow for significant compression while yielding large memory savings.

\begin{figure*}[t]
\centering
\includegraphics[width=1\linewidth]{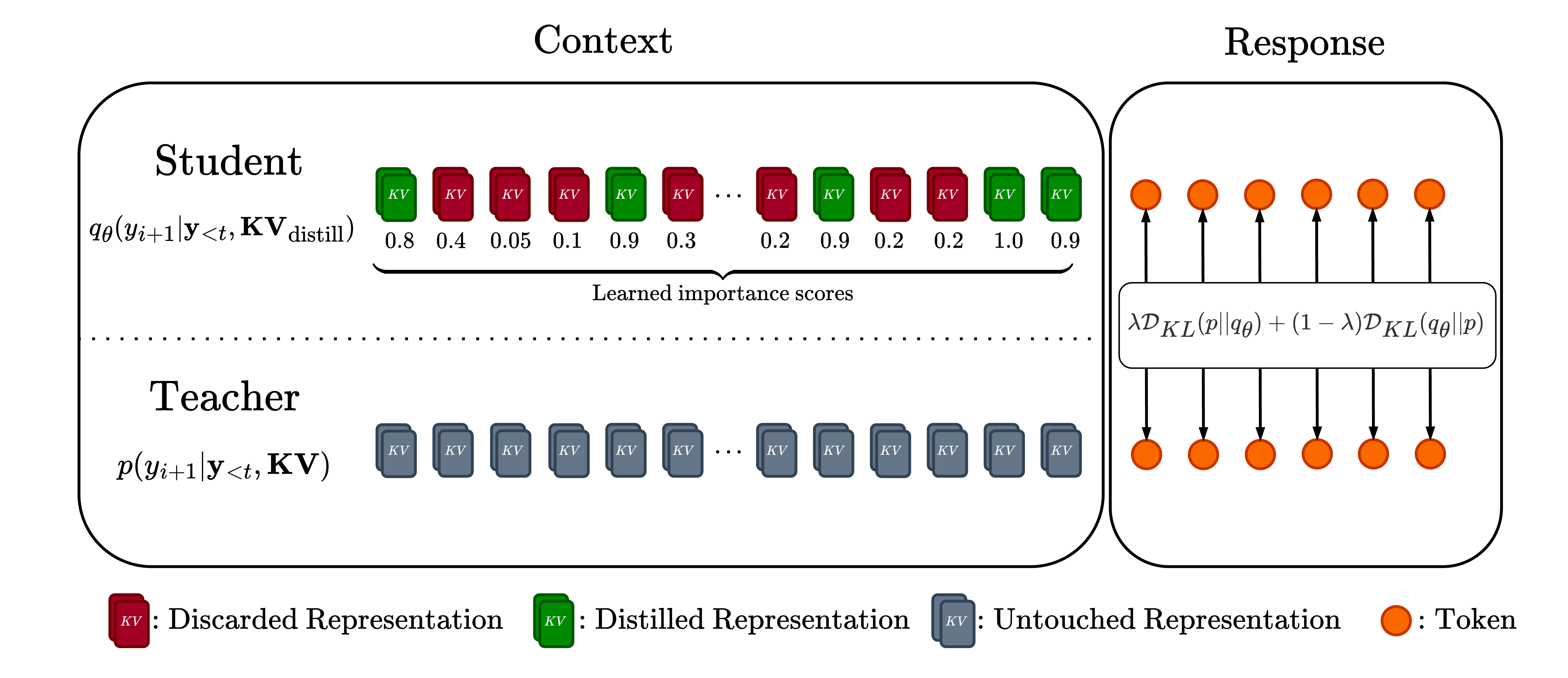}
\caption{
We subselect tokens from the \kv cache and distill into the smaller subset
}
\label{fig:overview}
\end{figure*}

In what follows, we suppose that a prompt to a LLM is composed of contextual text(s) followed by a question whose answer is dependent on the provided context. \kv compression can be divided into two paradigms: \textit{question-aware}, and \emph{question-independent}. In question-aware compression, we have access to the question that we need answered, and can compress the context with this in mind. In question-independent compression, we do not know what questions will be asked in the future. For instance, consider a scenario in which a fixed textual context will be used to respond to many questions; the goal of {question-independent} compression is to compress this context once for reuse across many question.  

Prior work in training-free context compression has primarily focused on which representations in the \kv cache to select for eviction, with excellent results \citep{zhang2023h2oheavyhitteroracleefficient,li2024snapkv}. In practice we observe that the performance of this selection procedure suffers greatly in the question-independent paradigm. Furthermore, we anticipate that there is room for performance improvements in general-purpose context compression when the model is trained to handle for compression. 

Prior work in trainable context compression have typically utilized a combination of cross-entropy and autoencoding objectives to pre-train general context compressors \citep{dodo24, Ge2023-rk, Rae2019-jo}, which are suitable for question-independent compression. These loss functions have led to significant performance loss at high compression rates.

In this work we design a general-purpose trainable context compression method for LLMs that outperforms prior methods in both the \emph{question-independent} and \emph{question-aware} paradigms. Our method, \method, accomplishes this, while also maintaining pretrained model capabilities, being suitable for long contexts, and having minimal performance penalty on downstream tasks. \method can support coherent, useful generation at compression ratios as high as 1000x.

To achieve this we train a scorer which retains the most important context tokens, while applying a parameter efficient adapter to conditionally modify important tokens' activations in-place. We further apply a token-level KL-type divergence to match the next-token prediction distributions, treating the compressed cache as a student, and the uncompressed cache as a teacher. \method only need be applied once to a fixed context, has zero overhead during auto-regressive decoding, and can compress arbitrary (sub)spans of a given context. We show improvements on several model families, considering extractive and abstractive tasks, with both short and long contexts, and at multiple model scales. \method is general purpose and has broad applicability to the LLM community.

\section{Background}

\newcommand{\ho}{$\mathsf{H_2}$\ }

\subsection{Key-Value Cache}
\label{sec:kvcachebackground}

Transformer-based language models (LMs) \citep{transformers17} use self-attention to aggregate context information and make predictions.
A decoder-only transformer LM \emph{autoregressively} predicts new tokens, and each step requires the LM to obtain the key and value states of all past tokens.
To avoid re-computing the \kv state of past tokens, most LM implementations  (e.g. \citet{hftransformers20}) cache the key and values states, in a structure called the \kv cache.
When making new predictions, self-attention is performed on query states of the new token and the \kv-cache, and the new token's key and value representations are appended to the \kv cache.
Because the \kv cache grows proportional to the number of tokens generated, maintaining the full KV cache in memory is a primary bottleneck when conditioning on large contexts.
The goal of this work is to alleviate this by \emph{compressing KV cache in the dimension of sequence length}, especially in the question-independent regime

\subsection{Related Work}
\label{sec:related_work}

Much prior work has tackled the problem of reducing the complexity of the self-attention mechanism itself. 
Previous work tries to sparsify the attention patterns\citep{Beltagy2020-dv,bigbird20}, use recurrence attention\citep{xlnet19}, or kernelize the attention matrix\citep{performer21}, but they require a considerable amount of further training.

Similar to our work, one line of work involves compressing the hidden states (KV cache) of past tokens into a shorter sequence of representations.
For example, some methods learns ``soft representations'' of context \citep{softprompt21}. \citet{gist23} compress particular prompts into much shorter ``gist tokens'', but do not attempt more general context compression. Furthermore, their method demonstrates poor generalizability, as performance does not scale with the number of gist tokens used. \citet{vcc23} propose to recognize and prioritize some important tokens (VIP tokens) during inference. Most relevant here, the following methods employ a similar idea of dynamically compressing the context prior to inference.

\subsection{Trainable Compression}

\citet{Ge2023-rk} design In-Context Autoencoder (\textsc{icae}) to compress long contexts for use in large language models (LLMs). \textsc{icae} consists of two main components: a learnable encoder and a fixed decoder. The encoder compresses the input context into a small number of memory slots. 
These memory slots are then used by the frozen LLMs (decoder) to reconstruct the context or respond to prompts.
\textsc{icae} is pretrained using autoencoding and language modeling objectives on a large pretraining corpus and further fine-tuned using instruction data to maintain instruction-tuning. 
However, there is still a gap in downstream task performance when using an \textsc{icae}-compressed context, compared to an uncompressed context, and the method falters under high compression ratios. 

\citet{dodo24} propose \textsc{dodo} to compress sub-select \kv activations to a set of ``nugget'' tokens, which grow proportionally with the length of context sequence. Their method is trained with auto-encoding or language modeling objectives. However, \textsc{dodo} models operate at a fixed compression ratio, require training both an encoder and decoder, and still show a large gap in downstream task performance when compared to an uncompressed context.

\subsection{Training-Free Compression} \citet{zhang2023h2oheavyhitteroracleefficient} propose \ho to reduce memory usage during generation. \ho identifies ``heavy-hitter'' tokens, which significantly influence attention scores during inference. Specifically, \ho calculates the accumulated attention for each key and retains the top-$k$ key-value pairs with the highest scores. In the question-aware setting, the accumulated attention scores include  scores from tokens in the question attending to the context. This effectively uses the question to scan for important details in the context. This allows \ho to maintain nearly uncompressed performance at moderate compression ratios, by focusing on tokens most relevant to the current question. However, performance still degrades when compression ratios exceed 20×.

In the question-independent paradigm, the \ho selection mechanism is applied solely to the context (as opposed to the context and question in the question-aware setting). We then allow the question to attend to only to this compressed context. We empirically observe that in the question-independent paradigm, \ho performance plummets drastically, highlighting the need for improved question-independent compression methods. Lastly, \ho offers no way to further improve compressive performance given prior domain knowledge. 

Similarly SnapKV \citep{li2024snapkv} uses the attentions of a window of recent tokens to determine which context tokens are ``heavy-hitters"; in the question-independent setting, this is undesirable, as the last tokens of a context may not necessarily provide additional information regarding attention patterns. In the question-independent paradigm we find that SnapKV performs similarly to \ho, so do not compare against it in the remainder here.

\section{Key-Value Distillation}
\newcommand{\lm}{\mathtt{LM}}
\newcommand{\compr}{\tilde{\vec{X}}}
\newcommand{\kl}{\mathcal{D}_\mathrm{KL}}


We consider a transformer-based language model \citep{transformers17}, denoted by $\lm$, that is defined on the vocabulary $\mathcal V$. The \method \ process is then: (1) a set of important tokens in the input context is determined; (2)  an adapted language model $\lm_\theta$ is used to encode the context into a \kv cache, and sub-select the aforementioned important tokens from the generated \kv cache; and (3) the unmodified $\lm$ conditions on the compressed \kv cache to auto-regressively generate it's output. 
\subsection{State Selection for Cache Compression}
\label{sec:subselection}

Let $\mathbf{c}=\{w_i\}_{i=1}^{N}$ represent a context consisting of $N$ tokens, where $w_i \in \mathcal V$ and $\vec c \in \mathcal V^N$.
In a typical scenario, $\lm$ predicts a sequence of new tokens, denoted by $\vec y$, conditioning on $\mathbf{c}$.
For example, $\mathbf{c}$ may be a prompt and $\lm$ generates $\vec y$ as a response.
Future token prediction draws on information from past tokens via \emph{attention} by having $\lm$ encode the context tokens into key and value hidden states $\vec X^{(K)}_{l},\vec X^{(V)}_{l}\in\mathbb {R}^{N\times d}$, which taken together form the \kv cache (Section \ref{sec:kvcachebackground}), where $d$ is the dimension of the transformer and $l$ is the layer of $\lm$.
We may drop the subscript $l$ and superscripts $^{(K)}$ and $^{(V)}$ and use $\vec X$ to generally denote the key/value states of transformers at any layer.

Transformers assume that $\vec X$ fully describes and represents the context $\vec c$.
However, attending to $\vec X$ can be inefficient when $\vec c$ is long.
Therefore, we further assume that retaining {a subset of key/value states is sufficient for approximating the next-token distribution conditioned on all key/value states}.
That is, we could \emph{retain} rows from $\vec X$ to form $\compr\in\mathbb R^{k\times d}$, where $k\le N$ is the number of selected rows.
We use a subset of the tokens' hidden states to represent the complete context, which is plausible because representations in $\compr$ are conditioned on the prior context.
Suppose we determine the $(i_1, \dots, i_k)$-th tokens are to be retained in layer $l$. We use a hard selection matrix $\vec S_l \in \{0,1\}^{k\times N}$ to derive (layer-specific) $\compr_l$ from $\vec X_l$ by
\begin{equation}
        \compr_l = \vec S_l \vec X_l, \quad  \vec S_l = [\vec{e}_{(i_1)}, \dots \vec ,\vec e_{(i_k)}],
\end{equation}
where $\vec e_{(i)}\in\{0,1\}^{N}$ is the $i$-th standard basis vector. Note that this formulation does not require that the same tokens be selected across each layer. 

The problem of determining which indices $(i_1, \dots, i_k)$ to retain still remains. We would like the subselection $\vec S$ to retain most of the context information given a fixed $k$. One possibility is to use a feedforward neural network to measure the importance of each token position:
\begin{align}
    \vec{s} = \mathtt{FFN}_\theta\left( \vec{X'}_\eta \right)
\end{align}
where $\theta$ is the parameters of the FFN, $\vec s \in \mathbb R^{N}$ and $\vec s_i$ indicate the ``importance score'' of the $i$-th token and $\vec X'_\eta$ indicates the hidden states at the $\eta$-th layer. The indices $i_{1:k}$ can then be derived by taking the tokens with the top-$k$ scores. We can control the percent of the \kv cache retained by scaling $k$ with the length of the context. In our we retain the same $i_{1:k}$ across all layers and take $\eta=6$.

The above selection procedure is rendered non-differentiable by the top-$k$ operator.
We may propagate gradients to the scorer by decaying the attention weights of tokens attending to $\compr$ inversely proportionally to their computed importance scores. 
More precisely, let $\vec{z} \in \mathbb R^{d}$ represent the hidden state of a single token attending to $\compr$ with unnormalized attention weights $\alpha$:
\begin{equation}
    \alpha = \left(\vec{z} \vec W^\text {Q}\right)\left({\compr^{(K)}}\right)^\top
\end{equation}
we decay $\alpha$ to produce scorer-informed attention weights $\alpha'$:
\begin{equation}
    \alpha' = \sigma(\vec s) \odot \alpha,
\end{equation}
where $\odot$ denotes the element-wise (Hadamard) product and $\sigma$ the sigmoid function. 
We note that the above formulation is one of many possible scoring functions that can be used with \method, that could be learnable or parameter-free, and could potentially have layer-wise specificity. We leave exploring different scoring functions as future work.

\subsection{Architecture}
\label{sec:architecture}

\begin{figure}[t]
    \centering
    \includegraphics[width=0.9\linewidth]{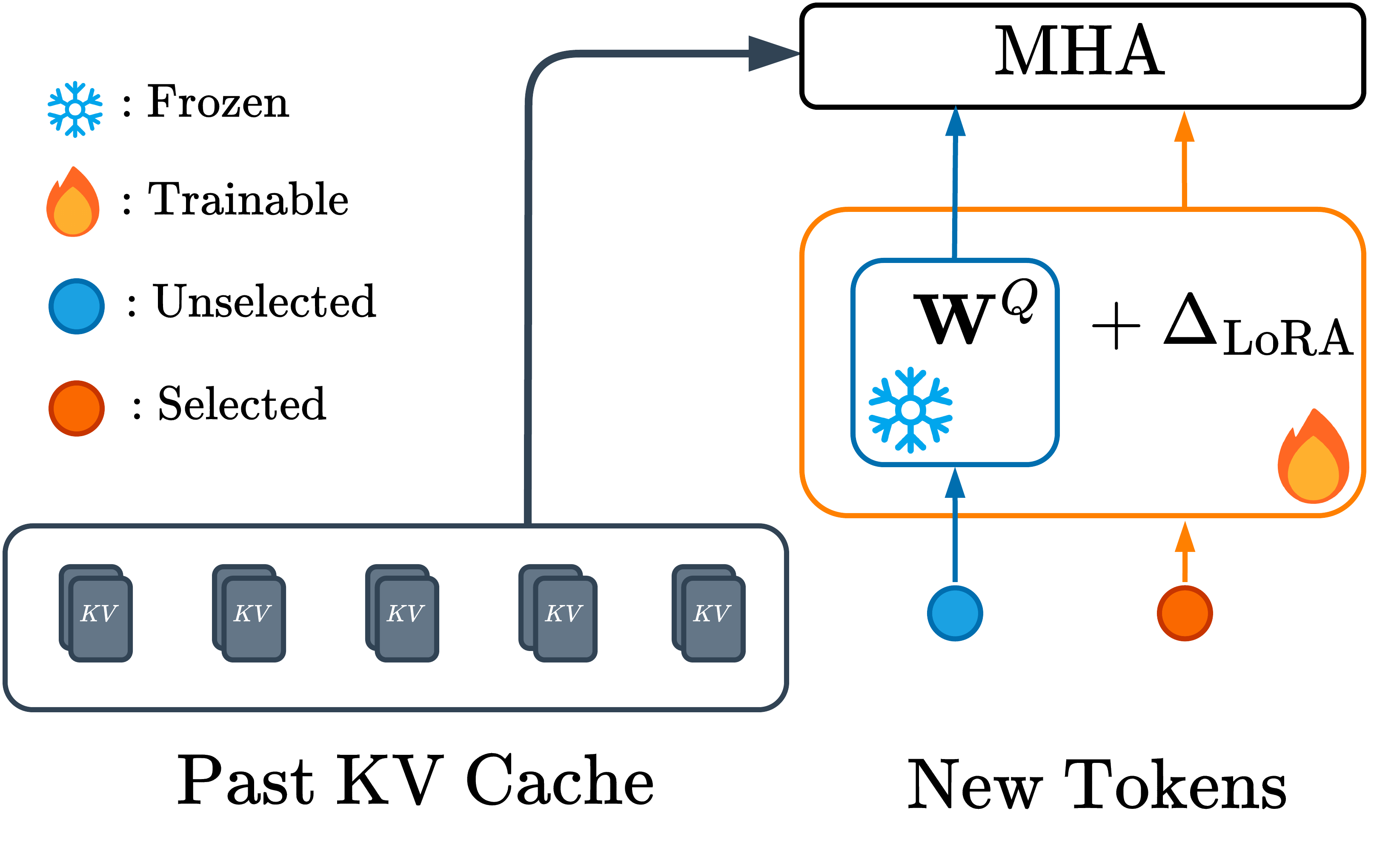}
    \caption{Selected tokens are routed to trainable, LoRA-adapted $\vec W^\text{Q}$ and $\vec W^\text{O}$ matrices ($\vec W^\text{O}$ is omitted in this figure); all other tokens pass through the original (frozen) model parameters.}
     \label{fig:arch}
\end{figure}
After  sub-selecting important token indices, we pass the context $\vec c$ through a modified $\lm_\theta$ that uses \emph{conditional computation} to condense the context into $\compr$. This allows for the representations of important tokens to be ``packed" with information from unselected tokens, and is strictly more expressive than only subselection. We instantiate $\lm_\theta$ with LoRA adaptors \citep{Hu2021-ya} to minimize the number of trainable parameters.

More importantly, within $\lm_\theta$, the subselected tokens are routed to trainable $\vec W^\text{Q},\vec W^\text{O}$ matrices, where  $\vec W^\text{Q},\vec W^\text{O}$ are the query and output matrices of transformers, while discarded tokens are routed to the original (frozen) matrices, as shown in Figure \ref{fig:arch}.
This has the effect of informing $\lm_\theta$ as to which tokens are selected, allowing for specialized aggregation of the value representations for  selected tokens. This method of informing $\lm_\theta$ has minimal overhead (the LoRA matrices account for under 500MB of GPU memory for a 27B parameter model), and only a single set parameters must be maintained in memory.

We anticipate that other architectural forms could make \method effective. However, we find that applying conditional computation to inform $\lm_\theta$ of selected tokens is  important to the performance of \method. We find that some methods of informing the model of selected tokens, such as by adding a trainable embedding to these tokens, do not work well (see Appendix~\ref{sec:ablation}). The particular architecture chosen has the advantage of lower memory usage during training, and provides excellent performance. We leave the task of finding even more efficient architectures to future work.
\subsection{Objective Function}\label{sec:objective}
After generating compressed cache $\compr$, we aim to match the output of $\lm$ when conditioned on $\compr$ to the output of $\lm$ when conditioned on $\mathbf{X}$.
Previous compression methods \citep{Ge2023-rk,nugget23} rely on the autoencoding objective to pretrain $\lm_\theta$. 
However, given that $\lm$ predicts future tokens during inference, there is a discrepancy in pretraining and downstream usage, which could result in performance loss. Instead we propose matching the next-token probability distribution of tokens conditioned on $\vec X$ and $\compr$. Consider a generative language model that predicts the next token $\vec y_t$ conditioned on the past tokens $\vec y_{<t}$ and a fixed context $\vec c$ that is represented by either $\vec X$ or $\compr$.
We would like to minimize the difference between their next-token distributions, i.e. 
$p\left(\vec y_t\mid \vec{y}_{<t}, \vec X\right)$ and $q_\theta\left(\vec y_t\mid \vec{y}_{<t}, \compr\right)$.
Let $q_\theta$ indicate the distribution that conditions on the distilled KV cache $\compr$.
Also note that the only learnable parameters in this formulation arise from \emph{encoding} $\compr$; during auto-regressive generation we use the original frozen parameters of $\lm$.

Given probability distributions $p, q_\theta$, we use the forward and reverse KL divergences to measure their similarity.
With simplified notations we have:
\begin{multline}
        \kl(p \| q_\theta) = \mathbb E_{y \sim p(\cdot)} \left[\log\left(\frac{p(y)}{{q_\theta}(y)}\right)\right]\\ \kl\left(q_\theta \| p\right) = \mathbb E_{y \sim q_\theta(\cdot)} \left[\log\left(\frac{{q_\theta}(y)}{p(y)}\right)\right]
    \label{eq:kls}
\end{multline}

The mode-seeking and mean-seeking behavior of the reverse- and forward- KL divergences respectively is well known. To incorporate both behaviors into the objective, we sum the forward and reverse divergences:
\begin{equation}
    \mathcal L(\theta) = \lambda \cdot \kl(p \| q_\theta) + (1-\lambda) \cdot \kl(q_\theta \| p),
    \label{eq:klsum}
\end{equation}
where a hyperparameter $\lambda$ controls the balance between forward and reverse KL divergence.

Given both $p$ and $q_\theta$ are categorical distribution, both KL divergences in \cref{eq:kls} can be analytically solved. Tthe L$1$-norm of the gradient of the reverse divergence dominates nearly everywhere. 
As such we propose scaling the forward and reverse terms by having $\lambda > 0.5$ in \cref{eq:klsum}. The benefit of $\lambda$ is confirmed with the ablations in Appendix~\ref{sec:ablation}.

\section{Experiments}

To assess the efficacy of \method, we conduct experiments on \textsc{llama-2 7b}, \textsc{llama-3 8b}, \textsc{mistral 7b},\textsc{gemma-2 9b} and \textsc{gemma-2 27b}. In all cases we use the instruction-tuned model. A \method model is obtained by distilling on a large corpus to obtain strong general-purpose context compressors.

\begin{figure*}[t]
\centering
\includegraphics[width=0.95\linewidth]{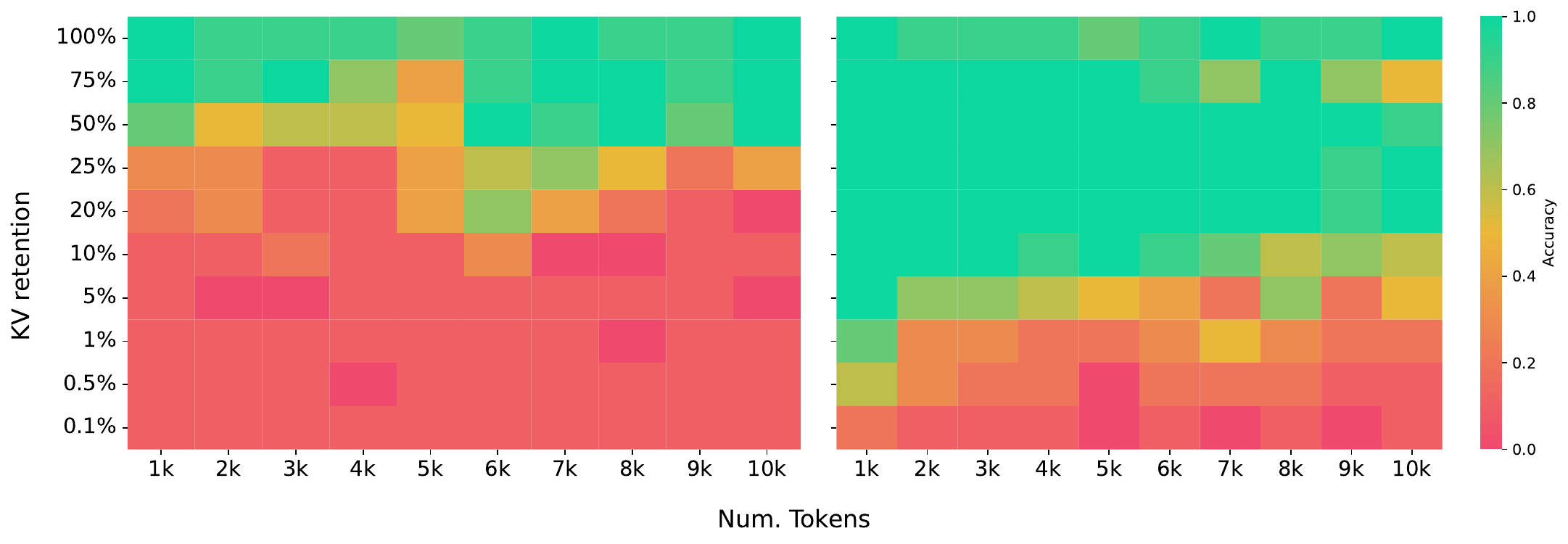}
\caption{Needle-in-a-Haystack results; The $x$-axis shows the length
of the document, the $y$-axis indicates the compression ratio applied, and the color the accuracy of retrieval under those settings averaged across different locations in the document. \textbf{Left:} $\mathsf{H_2I} $. \textbf{Right:} \method.}
\label{fig:needle}
\end{figure*}

\noindent\textbf{Data}\quad We curate a large instruction dataset from Self-Instruct, P3, LongAlpaca, and Super-Natural Instructions \citep{cerebras2023slimpajama,selfinstruct, sanh2021multitask,long-alpaca,wang-etal-2022-super}. Training instances are split into $(\textrm{Context}, \textrm{Instruction}, \textrm{Answer})$ triples.  In cases where the context is sufficiently long (more than 1536 tokens), we pad to a multiple of 1536 and fold the context to a batch of $N \times 1536$ instances, compress the resulting $\kv$ cache, and then unfold the cache. Empirically, we observe little performance degradation when applying folding during pretraining, while allowing the model to see longer examples. We also always leave the first few ($<10$) tokens of the context uncompressed, as we find that retaining them improves performance; this is not a new observation, see \citet{han-etal-2024-lm} and \citet{xiao2023streamingllm}.

\noindent\textbf{Training}\quad The general training procedure is as follows: (1) pass the $(\textrm{Context}, \textrm{Instruction}, \textrm{Answer})$ triple through the original model to obtain target logits, (2) apply the \method architecture to obtain logits conditioned on the compressed cache (we compress the context, and leave the instruction uncompressed), (3) apply Equation \ref{eq:klsum} between the obtained and target logits. 

We use rank-stabilized LoRA on the $Q,K,V,O$ matrices with $r=128$ to train $\lm_\theta$ \citep{Hu2021-ya,Kalajdzievski2023-pv}. Note that the $K,V$ are trainable for all tokens, not just selected tokens. The behavior of the $Q,O$ adapters is discussed in Section \ref{sec:architecture}. Optimization is done using Deepspeed Stage 2, and the AdamW optimizer \citep{10.1145/3394486.3406703}. During pretraining, we sample \kv retention fractions between $0.1\textrm{-}80\%$. As such, all \method models support arbitrary retention rates.  All models are distilled on a cluster of 8 NVIDIA A100 80GB GPUs. All models except \textsc{gemma 27b} converged within 3 days, while \textsc{gemma 27b} took 4 days. See Appendix \ref{sec:addtl_training} for further details.

\begin{table}[h]
  \centering
  \begin{tabular}{lcc}
    \toprule
    \textbf{Dataset} & \textbf{Average}  &  \textbf{Max}\\
    \midrule
    {SQuAD}     & 225           & 1k \\
    {QuALITY}     & 6k           &  9k \\
    {SQuALITY}     & 7k           & 11k \\
    {GovReport}     & 10k           & 71k \\
    \bottomrule
  \end{tabular}
    \caption{Evaluation dataset example length statistics in tokens under the \textsc{llama-3} tokenizer} \label{tab:stats}
\end{table}

\noindent\textbf{Evaluation}\quad In all datasets, we have a natural (\textrm{Context}, \textrm{Question}) pairing. We always compress the context and leave the question uncompressed. Evaluations are performed with greedy decoding. Summary statistics regarding the context length of evaluation dataset are provided in Table \ref{tab:stats}.

\noindent\textbf{Methods Tested}\quad We evaluate against \textsc{dodo} \cite{dodo24}, \textsc{icae} \cite{Ge2023-rk}, and \ho in both the question-aware ({$\mathsf{H_2A}$}) and question-independent ({$\mathsf{H_2I}$}) forms. Please refer to Sec~\ref{sec:related_work} for further description about the selected methods. To improve the performance of \ho, we also retain a set of sink tokens as described in \citet{xiao2023streamingllm}. 

\section{Results}
\subsection{Needle-In-a-Haystack}

\textbf{Motivation}\quad The Needle-in-a-Haystack test \citep{kamradt2023needle} evaluates a model's ability to accurately retrieve information from a sentence (a "needle") embedded within a large document (a "haystack"), in which a sentence is randomly positioned. In our experiments, the needle is a semantic one. In Figure~\ref{fig:needle}, we show the results of $\mathsf{H_2I}$ (left) and \method (right) at various compression ratios at different document lengths. The accuracy is computed across the placement of the needle within a document. Crucially, at compression time, the model does not know that a is needle being sought, nor that a needle is placed within the context.

\noindent\textbf{Results} We see that \method significantly outperforms $\mathsf{H_2I}$ at almost all compression ratios and document lengths. In particular, \method demonstrates  near-perfect accuracy even after removing $90\%$ of the \kv.

\subsection{Extractive Question Answering}

\textbf{Motivation}\quad SQuAD is an extractive question-answering task. We hypothesize that extractive tasks will suffer the largest performance loss under context compression. As such, we choose to use performance on SQuAD as a proxy for general-purpose compressive ability of a model. In all the following experiments, we choose the pretraining checkpoint with the best SQuAD performance for further experimentation. To assess accuracy we generate an answer conditioned on the compressed context, checking if the generated response is contained in the ground-truth answer.

\begin{table}[t]
\small
    \centering
  \begin{tabular}{lccc}
    \toprule
    \textbf{Model} & & \textbf{\% KV}& \textbf{0-Shot Acc.}\\
    \midrule
    \textsc{llama3}   & \textsc{base} & 100\% &$87.6 \pm .6$\%           \\
    \cmidrule(lr){2-4}
    & \cellcolor{LightCyan} \textsc{kvd} & \cellcolor{LightCyan}25\%&\cellcolor{LightCyan}$86.6 \pm .7$\%        \\
    & \cellcolor{LightCyan} \textsc{kvd}& \cellcolor{LightCyan}20\%&\cellcolor{LightCyan}$86.0 \pm .7$\%        \\
    \cmidrule(lr){2-4}
    & {$\mathsf{H_2A}$}   & 25\%& $84.0 \pm .7$\%        \\
    & {$\mathsf{H_2A}$}  & 20\%& $83.0 \pm .7$\%        \\
    & {$\mathsf{H_2I}$}    & 25\%& $56.6 \pm .9$\%        \\
    & {$\mathsf{H_2I}$}   & 20\%& $51.7 \pm 1$\%        \\
    \cmidrule(lr){2-4}
    & $\textsc{dodo}$    & 20\%& $73.3 \pm .8$\%        \\
    \midrule
    \textsc{llama2 7b}   & \textsc{base} & 100\% &$82.5 \pm .7$\%           \\
    \cmidrule(lr){2-4}
    & \cellcolor{LightCyan}\textsc{kvd}&\cellcolor{LightCyan} 25\%& \cellcolor{LightCyan}$79.1\pm .8$\%        \\
    & \cellcolor{LightCyan}\textsc{kvd}&\cellcolor{LightCyan} 20\%&\cellcolor{LightCyan}$77.6\pm .8$\%        \\
    \cmidrule(lr){2-4}
    & $\mathsf{H_2A}$    & 25\%& $77.9 \pm .7$\%        \\
    & $\mathsf{H_2A}$    & 20\%& $76.7 \pm .7$\%        \\
    & $\mathsf{H_2I}$    & 25\%& $55.2 \pm .9$\%        \\
    & $\mathsf{H_2I}$    & 20\%& $50.3 \pm 1$\%        \\
    \cmidrule(lr){2-4}
    & $\textsc{icae}$ & $57\%$    &$75.0 \pm .8$\%        \\
    \midrule
    \textsc{gemma 9b}   & \textsc{base} & 100\% & $85.15 \pm .7$\%           \\
    \cmidrule(lr){2-4}
    & \cellcolor{LightCyan}\textsc{kvd} & \cellcolor{LightCyan}25\%&\cellcolor{LightCyan}$84.55 \pm .7$\%        \\
    & \cellcolor{LightCyan}\textsc{kvd}& \cellcolor{LightCyan}20\%&\cellcolor{LightCyan}$83.1 \pm .7$\%        \\
    \midrule
    \textsc{gemma 27b}  & \textsc{base} & 100\% & $85.3 \pm .8\%$           \\
    \cmidrule(lr){2-4}
    & \cellcolor{LightCyan}$\textsc{kvd}$& \cellcolor{LightCyan}25\%& \cellcolor{LightCyan}${83.1 \pm 1\%}$        \\
    & \cellcolor{LightCyan}$\textsc{kvd}$& \cellcolor{LightCyan}20\%& \cellcolor{LightCyan}${82.2 \pm 1\%}$        \\
    \midrule
    \textsc{mistral 7b}   & \textsc{base} & 100\% & $87.1 \pm .6$\%           \\
    \cmidrule(lr){2-4}
    & \cellcolor{LightCyan}\textsc{kvd}& \cellcolor{LightCyan}25\%&\cellcolor{LightCyan}$84.1 \pm .7$\%        \\
    & \cellcolor{LightCyan}\textsc{kvd}& \cellcolor{LightCyan}20\%&\cellcolor{LightCyan}$82.5 \pm .7$\%        \\
    \bottomrule
  \end{tabular}
    \caption{Zero-shot accuracy on SQuAD at selected \kv retention ratios.  
    }  \label{tab:squad}
\end{table}
\noindent\textbf{Results}\quad Table \ref{tab:squad} contains {SQuAD} accuracy results. We see that in all cases, \method models perform within a few percentage points of base models, even under a ``worst-case" task. Furthermore, \method models significantly outperform prior trainable methods (\textsc{icae}, \textsc{dodo}), even when retaining less of the \kv cache. \method models significantly outperform {$\mathsf{H_2I}$}, demonstrating the ability of the pretraining objective and architecture to create reusable compressed \kv representations. \method models also enjoy a slight improvement over {$\mathsf{H_2A}$} at similar compression ratios, demonstrating the effectiveness of its compression at capturing almost all salient information in the context, even without question-awareness.

When retaining under 20\% of \kv, we observe rapid declines in performance across all methods, indicating the difficulty of the task under high context compression. Lastly, we note that initial pretraining hyperparameters for all models were set based on initial experimentation with \textsc{llama-3} and SQuAD; as such, we anticipate that performance of most models can be improved with hyperparameter tuning during the pre-training process. 
 
\subsection{Long Context Question Answering}

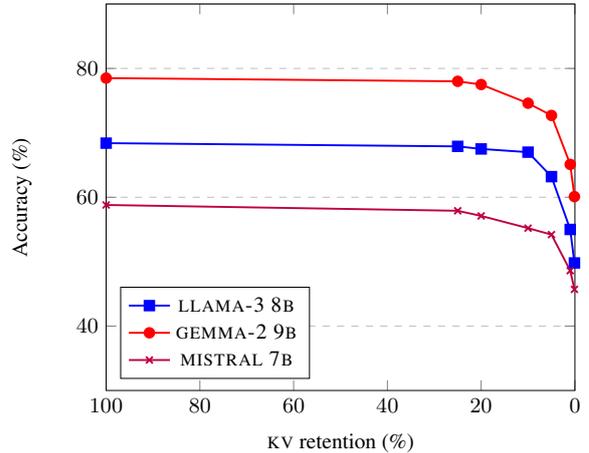
\begin{figure}[t]\small
\centering
\begin{tikzpicture}[scale=0.90]
\begin{axis}[
    xlabel={$\textsc{kv}$ retention (\%)},
    ylabel={Accuracy (\%)},
    xmin=0, xmax=100,
    ymin=30, ymax=90,
    x dir=reverse,
    ytick={0,20,40,60,80,100},
    legend pos=south west,
    ymajorgrids=true,
    grid style=dashed,
    every axis plot/.append style={thick},
]
\addplot[
    color=blue,
    mark=square*, 
    error bars/.cd,
    ]
    coordinates {
    (0.1, 49.8)+-(0,3.6)
    (1, 55.0)+-(0,3.6)
    (5, 63.2)+-(0,3.6)
    (10, 67.0)+-(0,3.6)
    (20, 67.5)+-(0,3.6)
    (25, 67.9)+-(0,3.6)
    (100, 68.4)+-(0,3.6)
    };
\addlegendentry{\textsc{llama-3 8b}}

\addplot[
    color=red,
    mark=*,
    error bars/.cd,
    ]
    coordinates {
    (0.1, 60.1)+-(0,3.6)
    (1, 65.1)+-(0,3.6)
    (5, 72.7)+-(0,3.6)
    (10, 74.6)+-(0,3.6)
    (20, 77.5)+-(0,3.6)
    (25, 78.0)+-(0,3.6)
    (100, 78.5)+-(0,3.6)
    };
    \addlegendentry{\textsc{gemma-2 9b}}

\addplot[
    color=purple,
    mark=x,
    error bars/.cd,
    ]
    coordinates {
    (0.1, 45.7)+-(0,3.6)
    (1, 48.6)+-(0,3.6)
    (5, 54.2)+-(0,3.6)
    (10, 55.2)+-(0,3.6)
    (20, 57.1)+-(0,3.6)
    (25, 57.9)+-(0,3.6)
    (100, 58.8)+-(0,3.6)
    };
    \addlegendentry{\textsc{mistral 7b}}   
\end{axis}
\end{tikzpicture}
\caption{QuALITY accuracy against compression.}
\label{fig:quality}
\end{figure}

\textbf{Motivation}\quad QuALITY is a long document multiple-choice question answering dataset that assesses reading comprehension. We use QuALITY to assess the decision making capabilities of models equipped with distilled contexts. To assess QuALITY accuracy, we use the same evaluation procedure used by \textsc{llama}-3 \citep{llama3modelcard}.

\noindent\textbf{Results}\quad \Cref{fig:quality} shows the experiment results on QuALITY, with data points at the following retention rates highlighted: $\{100\%, 25\%,$ $ 20\%, 10\%, 5\%, 1\%, 0.1\%\}$. We observe that \method performs similarly to the uncompressed cache, with only minor losses in performance at $10$x compression. Although not included in Figure \ref{fig:quality}, $0\%$ cache retention results in accuracy of 
$32.4\%$, $25.8\%$, and $24.4\%$ for the \textsc{llama-3}, \textsc{mistral}, and \textsc{gemma-2} models respectively, demonstrating the neccessity of the context for the task. Impressively, we see significant improvements over the random accuracy even when distilling to as few as 7 tokens from a 7k input passage; for example, on \textsc{llama-3} we observe only a $20\%$ drop in accuracy despite eliminating $99.9\%$ of the context.

\subsection{Long Context Abstractive Summarization}

\textbf{Motivation}\quad SQuALITY is a question-focused summarization dataset based on the same collection of long documents as the QuALITY benchmark. We use it to evaluate the abstractive summarization capabilities of models trained with distilled contexts. We compute the Rouge-L scores \citep{lin-2004-rouge} between the generated summaries and ground-truth answers, following the same evaluation protocol used by \textsc{llama}-3~\citep{llama3modelcard}. 

\begin{figure}[t]\small
\centering
\begin{tikzpicture}[scale=0.9]
\begin{axis}[
    xlabel={$\textsc{kv}$ retention (\%)},
    ylabel={\textsc{rouge-l}},
    x dir=reverse,
    xmin=0, xmax=100,
    legend pos=south west,
    ymajorgrids=true,
    grid style=dashed,
    every axis plot/.append style={thick},
]
\addplot[
    color=blue,
    mark=square*, 
    error bars/.cd,
    ]
    coordinates {
    (0.1, 14.36)
    (1, 14.87)
    (5, 15.39)
    (10, 15.50)
    (20, 15.49)
    (25, 15.48)
    (100, 15.47)
    };
\addlegendentry{\textsc{llama-3 8b}}

\addplot[
    color=red,
    mark=*,
    error bars/.cd,
    ]
    coordinates {
    (0.1, 13.89)
    (1, 14.34)
    (5, 14.41)
    (10, 14.65)
    (20, 14.82)
    (25, 14.85)
    (100, 14.84)
    };
    \addlegendentry{\textsc{gemma-2 9b}}

\addplot[
    color=purple,
    mark=x,
    error bars/.cd,
    ]
    coordinates {
    (0.1, 12.87)
    (1, 13.01)
    (5, 13.26)
    (10, 13.99)
    (20, 14.31)
    (25, 14.65)
    (100, 14.57)
    };
    \addlegendentry{\textsc{mistral 7b}}   

\end{axis}
\end{tikzpicture}

\caption{Rouge-L on SQuALITY}\label{fig:squality}
\end{figure}
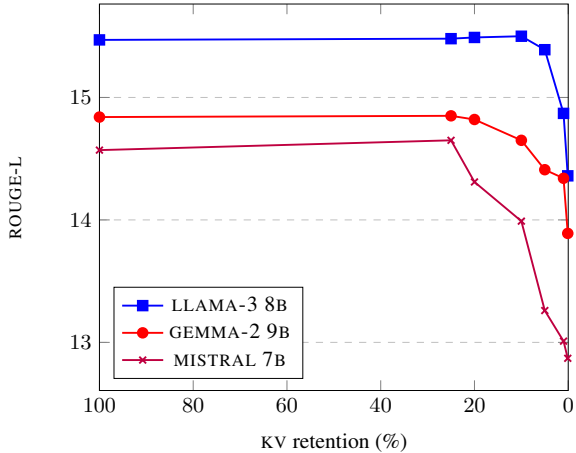

\noindent\textbf{Result}\quad Figure \ref{fig:squality} show Rouge-L performance on SQuALITY.
We observe that \textsc{kv-distill} models perform as well or better than uncompressed models when retaining more that 20\% of the \kv cache. When retaining under 20\%, we observe different performance falloff behaviors for different models; in particular, we observe that textsc{Llama-3} and \textsc{Gemma-2} have stable performance until 100x compression, at which point performance dips drastically. This difference in the behavior of the compression-performance trade-off could be attributed to the larger vocabulary sizes of \textsc{llama-3} and \textsc{gemma-2}, which allows the KL-loss to capture more fine-grained features of the output distribution during pretraining. These results demonstrate that \method can support very high compression ratios with minimal performance penalty on abstractive tasks.

\subsection{Finetuned Long Context Summarization}

\noindent\textbf{Motivation}\quad GovReport is a long document summarization dataset that consists (Report, Summary) pairs written by government research agencies. In contrast to the evaluations on QuALITY and SQuALITY (which are performed in a zero-shot fashion using the best pretraining checkpoint), we perform additional finetuning distillation with Equation \ref{eq:klsum} on the GovReport training set before evaluation. As with SQuALITY, we use GovReport to assess the abstractive summarization ability of models equipped with \method.

\begin{table}[t]
    \centering
    \small
  \begin{tabular}{lcccc}
    \toprule
    &&& \multicolumn{2}{c}{\textsc{kvd}} \\
    \cmidrule{4-5}
    \textbf{\kv retention} & $\mathsf{H_2A}$ & $\mathsf{H_2I}$ &\textbf{Zero Shot}  & \textbf{Finetune}  \\
    \midrule
    100\%     & 23.7 & 23.7 & 23.7           & 23.7 \\
    20\%     & 22.8 & 20.6 &\cellcolor{LightCyan}22.3         & \cellcolor{LightCyan}\textbf{23.5} \\
    10\%     & 22.4 & 18.6& \cellcolor{LightCyan}21.8           & \cellcolor{LightCyan}\textbf{23.3} \\
    5\%     & 21.9 & 18.5&\cellcolor{LightCyan}21.1           & \cellcolor{LightCyan}\textbf{23.2} \\
    1\%     & 21.1 & 18.3&\cellcolor{LightCyan}20.1           & \cellcolor{LightCyan}\textbf{22.8} \\
    \bottomrule
  \end{tabular}%
    \caption{\textsc{rouge-l} on GovReport summarization.}  \label{tab:govreport}
\end{table}

\noindent\textbf{Results}\quad Table \ref{tab:govreport} shows results for GovReport for both query-aware $\mathsf{H_2A}$ and query-independent $\mathsf{H_2I}$ paradigms, as well as \method prior to finetuning (zero-shot) and \method after finetuning on \textsc{llama}-3. We observe that \method and query-aware $\mathsf{H_2}$ perform close to each other on this evaluation in the zero-shot setting, while \method outperforms $\mathsf{H_2I}$ at all compression ratios. However, upon finetuning, we observe a practical improvement in performance with \method, with little degradation from uncompressed performance across all compression rates. In particular, we note the improvement in performance is greater at more severe compression ratios, confirming the utility of \method in supporting ultra-high compression ratios.

\section{Discussion and Conclusion}

We develop a method to reduce the memory requirements of long-context conditioned LM generation. Our method sub-selects tokens from the \kv cache, and applies a token-level KL-type loss between the output of the LM when conditioned on sub-selected tokens and when conditioned on the uncompressed cache. We evaluate our method on long-context extractive and abstractive tasks, and demonstrate improved performance over competing compression methods. We further demonstrate that continued training on domain-specific data can allow for use of compression ratios as high as 100x with negligible losses in performance. 

As part of this work we release distilled checkpoints across various model language families. These artifacts allow efficient text generation conditioned on significantly larger inputs than before, with much lower memory burden, and support compression ratios as high as 1000x. We anticipate these artifacts will be of great practical benefit, enabling exciting new applications and research directions in language processing.

\section{Limitations}

The time-consuming and stochastic nature of distilling a model means that it cannot be guaranteed that the process will work well across all model families. Furthermore, we are unsure as to the root cause of performance discrepancies between model architectures after distillation; this issue merits further research. Lastly, the 8k token context-capacity of \textsc{llama-3} limited many of our experiments, and is small by the standards of currently available language models; to address this, we will be releasing a \method \textsc{llama-3.1} model with a 128k token context capacity.

\noindent\textbf{Used Artifacts } In this work, we used the publicly released codes and checkpoints of LLAMA. Per the license attached to LLAMA, we agree not to re-distribute their parameters and limit the usage of the models for research purposes only

\bibliography{acl_latex}

\appendix

\section{Training \& Evaluation Details}\label{sec:addtl_training}
We train all \method models with the following parameters at bf16 precision on 8 NVIDIA A100s. Please see Table \ref{tab:hyper} for further details. All models had ~150M trainable parameters.
\begin{table}[h]
    \centering
    \begin{tabular}{c|c}
        \toprule
         \textbf{Hyperparameter} & \textbf{Value}\\
         \midrule
         Optimizer & AdamW\\
         Learning Rate & 5e-5 \\
         Batch Size & 32\\
         LoRA Rank & 128 \\
         $\lambda$ & 0.6\\
         $\eta$ &  6\\
         \bottomrule
    \end{tabular}\caption{Hyperparameters for training}\label{tab:hyper}
\end{table}
The QuALITY, SQuALITY, GovReport, and SQuAD evaluations are performed on the test set, if public, else results are reported on the development set. To measure SQuAD accuracy, we generate up to 128 tokens, normalize the output by stripping punctuation, and check if the correct answer is contained in the generated answer. For SQuALITY and QuALITY, we follow the evaluation procedure of \citet{llama3modelcard}. For GovReport, we prompt the model to summarize the report, and then greedily generate 630 tokens.

\section{Pretraining Objective Ablations}\label{sec:ablation}
We assess the necessity of both the forward and reverse terms in the loss by evaluating SQuAD performance on multiple different pre-training losses with varying $\lambda$ values in Equation \ref{eq:klsum}. In Table~\ref{tab:lambda} we observe that using either the pure forward or reverse divergences performs markedly worse than using a mixture of both. Furthermore, using solely the auto-encode + cross-entropy (used in \textsc{icae} and \textsc{dodo}), performs significantly worse than Equation \ref{eq:klsum}, demonstrating the significant benefits that the weighted distillation objective provide. We also note that replacing the routing mechanism with a learnable embedding (added to important tokens) does not perform well.

\begin{table}[h]
\small
  \centering
  \begin{tabular}{lc}
    \toprule
    \textbf{Loss} & \textbf{SQuAD Acc. (\%)} \\
    \midrule
    $\lambda= 1$    & 83.4\%    \\
    $\lambda= 0.6$    & 86.0\%    \\
    $\lambda= 0.4$    & 85.3\%    \\
    $\lambda= 0$    & 82.7\%    \\
    \midrule
    \textsc{ae + lm} & 79.1\%\\
    \textsc{no routing} & 67.4 \%\\
    \bottomrule
  \end{tabular}
    \caption{Effect of Pretraining Loss wrt \textsc{llama-3}.}\label{tab:lambda}
\end{table}


\section{Qualitative Analysis of Generated Text}
\label{sec:qual_appendix}
In Figure \ref{fig:example} we present representative summaries of the short story ``Galactic Ghost". \textsc{llama-3} was prompted to summarize the short story in a few sentences. As the \kv retention ratio decreases, we observe an increase in statements that are unfaithful to the original context. Particularly, at extremely high compression ratios, the model begins to hallucinate details, but impressively still maintains understanding of the general plot of the story. Note the varying degrees of \emph{semantic compression} in summaries generated under high compression ratios; qualitatively, we observe that models with severely compressed contexts generally understand the text, but resort to making vague statements. 

\begin{figure}[h]
\begin{mdframed}
    \centering
\begin{subcaptionblock}{\linewidth}
    \caption{100\% $\textsc{kv}$ retention}
    \vspace{3pt}
    \raggedright
    \small{The story follows Willard, a space explorer who is left alone after his friend and space mate, Dobbin, dies. Willard is haunted by the memory of Dobbin's final words, ``I saw the Ghost Ship," and becomes obsessed with finding the ship. He spends years searching for it, and eventually, he sees a strange, ghostly ship that seems to be heading towards him. As he gets closer, he realizes that it's not just a hallucination, but a real ship from Earth that's been traveling through space for centuries. In the end, Willard is taken aboard the ship, but learns that the ship and its crew are mere shells of their former selves, and that they are doomed to remain in space forever. Ultimately, Willard is forced to confront the reality that he'll never return to Earth.}
\end{subcaptionblock}
\begin{subcaptionblock}{\linewidth}
    \vspace{3pt}
    \caption{2.5\% $\textsc{kv}$ retention}
    \vspace{3pt}
    \raggedright
    \small{The story follows the journey of Willard, a space traveler who has been alone in space for years after the death of his friend and partner, Dobbin. Willard is struggling to come to terms with his isolation and the loss of his friend, and is haunted by the legend of the Ghost Ship, a mythical vessel that is said to roam the solar system, never to be seen again. One day, Willard spots a strange, rocket-shaped object approaching him, and as it draws closer, he realizes it is the \hl{Mary Lou}. The ship is now a mere shadow of its former self, and Willard is invited to board it, where he is greeted by a rescue team who have been sent to retrieve him. As he adjusts to life on the rescue ship, Willard comes to terms with his past and the reality of his situation, and ultimately accepts that he will never return to Earth. The story concludes with Willard's poignant realization that he and the crew of the rescue ship are the Ghosts, doomed to roam the solar system forever.}
\end{subcaptionblock}
\begin{subcaptionblock}{\linewidth}
    \vspace{3pt}
    \caption{.25\% $\textsc{kv}$ retention}
    \vspace{3pt}
    \raggedright
\small{The story follows the journey of Willard, a space explorer who has been traveling through the galaxy for years, searching for a \hl{new home for humanity}. He has lost his friend and companion, Dobbin, and is struggling to come to terms with his own mortality. One day, he discovers a faint light in the distance, which he believes to be a sign of life. As he approaches the light, he realizes it is a ghost ship, a ship that has been drifting through space for \hl{millions} of years, its crew and passengers \hl{long dead}. The ship is a relic of a bygone era, and Willard is forced to confront the reality that he will never be able to return to Earth, as the ship would pass through it, making it impossible to survive. In the end, Willard comes to accept his fate, realizing that he and the crew of the ghost ship are doomed to drift through space forever, a reminder of the transience of human existence.}
\end{subcaptionblock}
\end{mdframed}
\caption{\textsc{llama-3} was tasked with summarizing a 6k token short story at low \kv retention rates. Inaccuracies in the summary are highlighted yellow, and were determined by hand.}\label{fig:example}
\end{figure}%

\end{document}